\documentclass{sig-alternate-05-2015}

\usepackage{url}
\usepackage{amssymb}
\usepackage{amsmath}
\usepackage{graphicx}
\usepackage[table]{xcolor}
\usepackage{multirow}

\usepackage[utf8]{inputenc}
\usepackage[T2A,T1]{fontenc}

\newcommand{\mb}{\mathbf}
\newcolumntype{C}[1]{>{\centering\arraybackslash}m{#1}}

\definecolor{lightgray}{gray}{0.9}

\begin{document}


\CopyrightYear{2016}
\conferenceinfo{MM '16,}{October 15-19, 2016, Amsterdam, Netherlands}
\isbn{}\acmPrice{}
\doi{http://dx.doi.org/10.1145/2964284.2964309}

%

\title{Deep Cross Residual Learning for\\Multitask Visual Recognition}
\subtitle{}

\numberofauthors{2}
\author{
\alignauthor
Brendan Jou\\
       \affaddr{Electrical Engineering}\\
       \affaddr{Columbia University}\\
       \affaddr{New York, NY 10027}\\
       \email{bjou@ee.columbia.edu}
\alignauthor Shih-Fu Chang\\
       \affaddr{Electrical Engineering}\\
       \affaddr{Columbia University}\\
       \affaddr{New York, NY 10027}\\
       \email{sfchang@ee.columbia.edu}
}

\date{}

\maketitle

\begin{abstract}
Residual learning has recently surfaced as an effective means of constructing very deep neural networks for object recognition.
However, current incarnations of residual networks do not allow for the modeling and integration of complex relations between closely coupled recognition tasks or across domains.
Such problems are often encountered in multimedia applications involving large-scale content recognition.
We propose a novel extension of residual learning for deep networks that enables intuitive learning across multiple related tasks using cross-connections called cross-residuals.
These cross-residuals connections can be viewed as a form of in-network regularization and enables greater network generalization.
We show how cross-residual learning (CRL) can be integrated in multitask networks to jointly train and detect visual concepts across several tasks.
We present a single multitask cross-residual network with >40\% less parameters that is able to achieve competitive, or even better, detection performance on a visual sentiment concept detection problem normally requiring multiple specialized single-task networks.
The resulting multitask cross-residual network also achieves better detection performance by about 10.4\% over a standard multitask residual network without cross-residuals with even a small amount of cross-task weighting.
\end{abstract}

%
%


%
%

%
%



\section{Introduction}
\label{sec:intro}

\begin{figure}[t]
  \centering
  \includegraphics[width=3.25in]{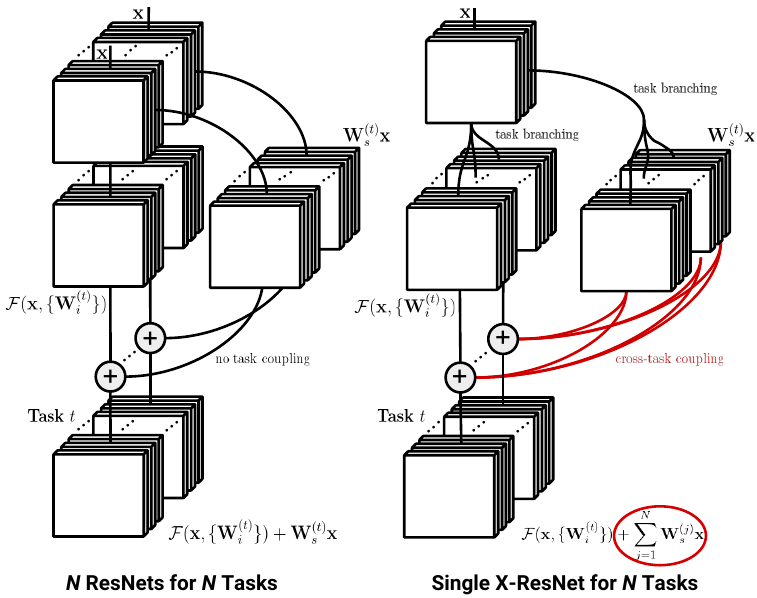}
  \caption{Feature Map Illustration of Residual Network (ResNet) and Cross-residual Network (X-ResNet) Layers. X-ResNet extends ResNet to enable structures like multitask networks where a single network can jointly perform multiple related tasks, as often encountered in multimedia applications, instead of requiring one network per task. Our network uses cross-task connections indicated in red to simultaneously enable specialization per task and overall generalization.}
  \label{fig:overview}
\end{figure}

In concept detection, leveraging the complex relationships between learning tasks remains an open challenge in the construction of many multimedia systems.
While some recent approaches have begun to model these relationships in deep architectures \cite{deng_2014,wu_2014}, still many multimedia solutions tend to have multiple parts that specialize rather than a more versatile, general solution that leverages cross-task dependencies.
As an illustration, visual sentiment prediction is a rising topic of interest in multimedia and vision.
In \cite{borth_2013}, a semantic construct called adjective-noun pairs (ANPs) was proposed wherein there are visual concept pairs like `happy girl', `misty woods' and `good food'.
These semantic concepts serve as a bridge between vision-based tasks that are focused on object (or ``noun'') recognition and affective computing tasks that are focused on qualifying the affective capacity or strength of multimedia, e.g.~through the ``adjective'' in the ANP.
However, even though the tasks of object recognition, affect prediction and ANP detection all have some relation to each other, the construction of classifiers for each is treated independently.
In this work, we propose a novel method for jointly learning and generalizing across tasks which can be easily and very efficiently integrated into a deep residual network and as a proof-of-concept show how it can be used for visual sentiment concept detection.

\begin{figure}[t]
  \centering
  \includegraphics[width=3.28in]{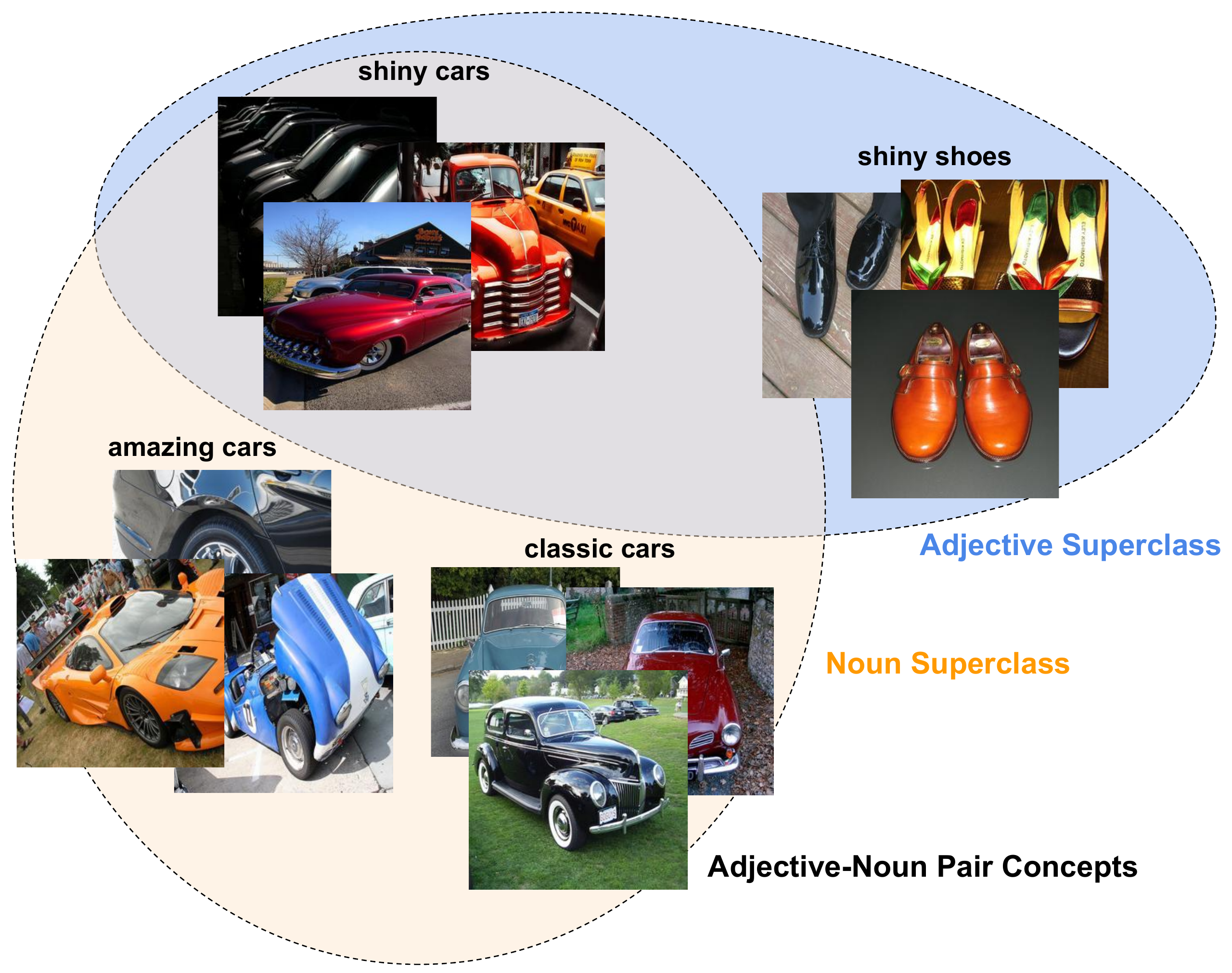}
  \caption{Example of related visual concept detection tasks that directly benefit from our proposed cross-residual learning (CRL). Adjective-noun pairs can be superclassed by their noun or adjective components. Exploitable visual and semantic similarities exist within (intra-relatedness) as well as between superclasses (inter-relatedness).}
  \label{fig:anp_relatedness}
  \vspace{-0.3cm}
\end{figure}

To understand how ``relatedness'' is both important and applicable to visual concept detection, consider several example images and concepts from \cite{borth_2013} in Figure \ref{fig:anp_relatedness}.
In the example, we observe that the ANP `shiny cars' can be superclassed by both the `shiny' adjective category and `cars' noun category.
Within the `shiny' adjective category, there are other concepts like `shiny shoes' that bear both semantic and visual similarities to the `shiny cars' ANP, e.g. gloss or surface luster.
This \emph{intra-relatedness} also exists within the noun superclass which includes ANPs like `amazing cars' and `classic cars'.
In addition to relatedness within the same (super)class, we observe that there are visual similarities also present between classes of different superclasses, e.g.~`classic cars' and `shiny shoes'.
This \emph{inter-relatedness} between (super)classes illustrates how in settings like concept detection, classifiers can benefit from exploiting representational similarities across related tasks.
Both of these senses of \emph{relatedness} show that visual representations across related tasks can be shared to a degree.
We develop a multitask learning problem for visual concept detection to illustrate a setting in which our proposed method can be applied.
We design a deep neural network with a stack of shared low-level representations and then higher level representations that both specialize and mix information across related tasks during learning.
We then show how such a multitask network architecture with cross-task exchanges can be used to simultaneously learn classifiers to detect adjective, noun and adjective-noun pair visual concepts.

In \cite{he_2016}, residual learning is proposed as an approach for enabling much deeper networks while addressing the \emph{degradation} problem where very deep networks have a tendency to \emph{underfit} compared to shallower counterpart networks \cite{srivastava_2015}.
In residual learning, an identity mapping through the use of shortcut connections \cite{raiko_2012} is proposed where an underlying mapping $\mathcal{H}(\mb x) = \mathcal{F}(\mb x) + \mb x$ is learned given that $\mathcal{F}(\mb x) = \mathcal{H}(\mb x) - \mb x$ represents another mapping fit by several stacked layers.
One interpretation is that $\mathcal{F}(\mb x)$ represents a noise term and the model is fitting the input plus some additive nonlinear noise.
Thus, if we were performing reconstruction, a trivial solution to the residual learning problem is that an identity mapping is optimal, i.e.~$\mathcal{F}(\mb x) = 0$.
However, in \cite{he_2016}, it is argued that optimization software may actually have difficulty with approximating identity mappings with a stack of nonlinear layers, and also that for prediction problems, it is unlikely that the strict identity is optimal.
They also argue that fitting residual mappings can enable deeper networks given the information boost achieved via the shortcut connection and thus reduces the likelihood of model degradation.
Our work extends residual learning \cite{he_2016} to also integrate information from other related tasks enabling cross-task representations.
Specifically, we hypothesize and experimentally show that reference components from correlated tasks can be synergistically fused in a residual deep learning network for \emph{cross-residual learning}.

Our contributions include
(1) the proposal of a novel extension of residual learning \cite{he_2016} using cross-connections for coupling multiple related tasks in a setting called cross-residual learning (CRL),
(2) the development of a multitask network with a fan-out architecture using cross-residual layers, and
(3) an evaluation of cross-residual networks on a multitask visual sentiment concept detection problem yielding a single network with very competitive or even better accuracy compared to individual networks on three classification tasks (noun, adjective, and adjective-noun pair detection) but uses >40\% less model memory via branching, while also outperforming the predictive performance of a standard multitask configuration without cross-residuals by about 10.4\%.

\section{Related Work}
\label{sec:relatedwork}

Our work broadly intersects three major lines of research areas: transfer learning, deep neural architectures for vision, and affective computing.
In traditional data mining and machine learning tasks, we often seek to statistically model a collection of labeled or unlabeled data and apply them to other collections.
In general, the distributions of these sets of data collections are assumed to be the \emph{same}.
In transfer learning \cite{pan_2010}, the domain, tasks and distributions are allowed to be \emph{different} in both training/source and testing/target.
In this work, we specifically focus on a subset of transfer learning problems that assume some \emph{relatedness} between these collections.
Specifically, in multimedia and vision contexts, \emph{relatedness} may refer to settings where groups of tasks have semantic correlation, e.g.~classifying dog breeds and bird species, or visual similarity, e.g.~jointly classifying and reconstructing objects, and is often referred to as multitask learning \cite{caruana_1997}.
Likewise, relatedness may also refer to the same source task but applied in different domains, e.g.~classifying clothing style across cultures, and is sometimes called cross-domain learning \cite{jiang_2008} or domain transfer/adaptation \cite{glorot_2011,jiang_2009}.
Nonetheless, the hypothesis of explicitly learning from related tasks is that we can learn more generalized representations with minimal performance cost or in some cases, leading to gains from learning jointly.

Multitask networks are recently becoming a popular approach to multitask learning, riding on successes in deep neural networks.
One early work in \cite{collobert_2008} showed how a single network could be trained to solve multiple natural language processing tasks simultaneously like part-of-speech tagging, named entity recognition, etc.
Multitask networks have since proven effective for automated drug discovery \cite{dahl_2014,ramsundar_2015}, query classification and retrieval \cite{liu_2015}, and semantic segmentation \cite{dai_2016}.
Recently, \cite{ghifary_2015} proposed multitask auto-encoders for generalizing object detectors across domains; and in \cite{luong_2016}, multitask sequence-to-sequence learning is proposed for text translation.
Also, architectures like \cite{ramus_2015,yim_2015} can be categorized as multitask networks since they reconstruct and classify simultaneously.
Unlike other multitask networks but similar to ladder networks \cite{ramus_2015}, instead of a single branching point in our network that creates forked paths to only specialize to individual tasks, we continue mixing information even after branching via our cross-skip connections.

Whereas multitask learning can generally be understood as a fan-out approach where a (usually, single) shared representation is learned to solve multiple tasks, an analogous complement is a fan-in approach where multiple either features or decision scores are fused together to solve a single-task.
For example, graph diffusion can be used smooth decision scores for leveraging intra-relatedness between categories \cite{jiang_2009}.
In \cite{deng_2014}, instead of an undirected graph, explicit directed edges were used to model class relationships like exclusion and subsumption.
And with some semblance to our work, in \cite{wu_2014}, a multimodal neural network structure is developed where inter-class (but still intra-task) relationships are integrated as an explicit regularizer.
Although inspired from multitask learning, the network design in \cite{wu_2014} still operates in a single-task context as there is only a single output network head.
Additionally, because the network integrates multiple input feature towers, the overall memory and training burden of the image-to-decision pipeline is much greater than that of a fan-out network alternative.

\begin{figure}[t]
  \centering
  \includegraphics[width=3in]{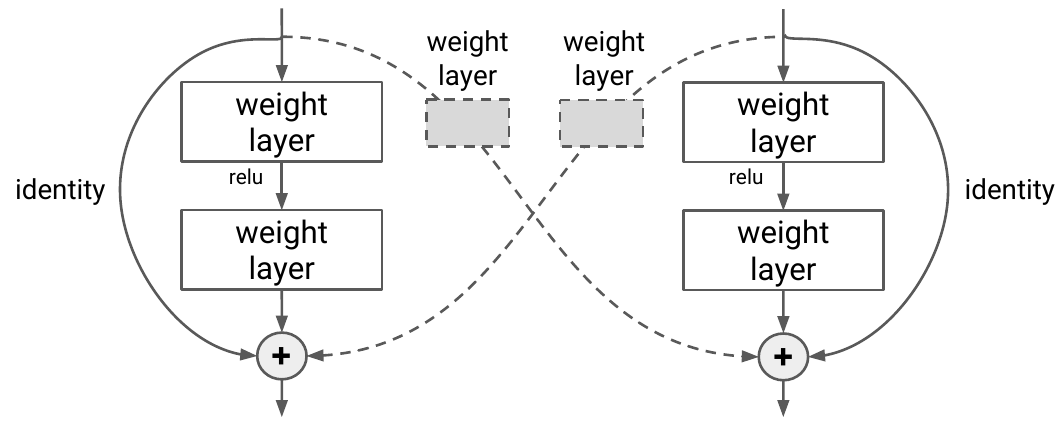}
  \caption{Cross-residual Building Block (with two tasks). Cross-residual weight layers and cross-skip connections are dashed and allow for network-level flexibility for task specialization.}
  \label{fig:xresidual}
\end{figure}

Since we use visual sentiment concept detection to illustrate the efficacy of our proposed cross-residual learning approach, it is worth also briefly noting several advances in visual affect.
In visual affective computing, a longstanding goal is to bridge the \emph{affective gap}, a conceptual disconnect between low-level multimedia features and high-level affective states like emotions or sentiment.
In \cite{yanulevskaya_2008}, a codebook over local color histogram and Gabor features were proposed for image-based sentiment prediction; and in \cite{machajdik_2010}, psychology and art theory inspired features were proposed.
Again, in \cite{borth_2013}, adjective-noun pairs were proposed as a mid-level semantic construct and an ontology was mined from a popular social multimedia platform using psychology-grounded seed queries \cite{plutchik_1980}.
Other problems related to affect detection include quality assessment \cite{ke_2006}, memorability \cite{isola_2011}, interestingness \cite{gygli_2013} and popularity \cite{khosla_2014}.
In this work, we develop a single deep multitask cross-residual network able to simultaneously predict noun, adjective and adjective-noun visual concepts.

\section{Cross Residual Learning}
\label{sec:xresidual}

Given an input $\mb x$ and output $\mb y$ vector to a residual learning layer and the mapping function $\mathcal{F}(\mb x, \{\mb W_i\})$ to fit, where for vision problems this might represent, for example, a stack of convolutional operations with batch normalization \cite{ioffe_2015} and ReLU activation \cite{nair_2010}, we have the following in residual learning \cite{he_2016}:
\begin{equation}
  \mb y = \mathcal{F}(\mb x, \{\mb W_i\}) + \mb W_s\mb x,
  \label{eq:residual}
\end{equation}
where $\mb W_s$ is an optional linear projection, but required when matching dimensions, on the shortcut connection.
For identity shortcut connections, $\mb W_s = \mb I$.

Here, we propose a simple and efficient extension of \cite{he_2016} when fitting across multiple related learning tasks which we refer to as \emph{cross-residual learning} (CRL).
Given a task $t$ and $N-1$ other related tasks, we define the task output of the cross-residual module as:
\begin{equation}
  \mb y^{(t)} = \mathcal{F}(\mb x, \{\mb W_i^{(t)}\}) + \sum_{j=1}^N \mb W_s^{(j)}\mb x,
  \label{eq:xresidual}
\end{equation}
where the superscript $(\cdot)$ indexes the target task and a normalization factor is omitted for simplicity and can be lumped with the shortcut weights $\mb W_s^{(j)}$.
As also illustrated in Figure \ref{fig:xresidual}, the other target tasks additively contribute to the current target task $t$ by $\sum_{j\neq t}\mb W_s^{(j)}\mb x$.
The cross-residual contributions can also more generally be stacks of operations $\mathcal{C}(\mb x,\{\mb W_{s,m}^{(j)}\})$, but here, we only illustrate the simple weighted once case $\mb W_s^{(j)}\mb x$.

\vspace{1.6mm}
\noindent \textbf{``Early'' Regularization Interpretation.}
In optimization, when minimizing a loss $\mathcal{L}(f(\mb x), \mb y)$, we often add a regularization term $\mathcal{R}(f(\mb x))$ to constrain the ``badness'' of the solution, factor in assumptions of our system, and reduce overfitting.
For example, in solving deep networks, the squared 2-norm is a common choice to penalize large parameter values and smooth network mappings.
Cross-residual units can be viewed as a way of regularizing the solution of a specific task by other related tasks, i.e.~we do not want the learned mapping $\mathcal{F}(\mb x, \{\mb W_i^{(t)}\})$ to be too far from a weighted combination of task-specialized transformations of the input $\sum_j \mb W_s^{(j)}\mb x$.
For example, when learning to visually recognize species of birds, we may want to introduce regularization to ensure the mapping fit is not too far from the separate, but related task of recognizing types of mammals.
While such a regularization usually takes place in the loss layer of a neural network, using cross-residual layers we can introduce this task conditioning ``earlier'' in the network and also stack them for additional information mixing.
Unlike typical ``regularization,'' a cross-residual layer introduces regularization by biasing at the layer-level, i.e. with respect to a given task’s residual rather than with respect to the penultimate loss.
Cross-residual layers thus serve as a type of in-network regularization somewhat similar to dropout \cite{srivastava_2014}, though with less stochasticity.

\vspace{1.6mm}
\noindent \textbf{Connection to Highway Networks} \cite{srivastava_2015} \textbf{ \& LSTM } \cite{hochreiter_1997}\textbf{.}
As also discussed in \cite{he_2016}, residual networks can be seen as highway networks \cite{srivastava_2015} that do not have transform or carry gates.
In highway networks, an output highway layer is defined as
\begin{equation}
  \mb y = \mathcal{H}(\mb x, \mb W_\mathcal{H}) \mathcal{T}(\mb x, \mb W_\mathcal{T}) + \mb x \cdot \mathcal{C}(\mb x, \mb W_\mathcal{C}),
  \label{eq:highway}
\end{equation}
where $\mathcal{T}$ and $\mathcal{C}$ are the transform and carry gates, respectively.
Clearly, when both gates are on, this is precisely the same as a residual layer.
By extension, a cross-residual layer can be thought of as an ungated highway layer with multiple ``highways'' merging onto the same information path.
Cross-residual weighting layers then are carry gates which govern the amount of cross-task pollination.

Similarly, it has been argued (though somewhat reductionist) that residual layers can also be viewed as a feed-forward long short-term memory (LSTM) \cite{hochreiter_1997} units without gates.
Specifically, consider the LSTM version from \cite{gers_2002}:
\begin{equation}
 \left.\arraycolsep=1.4pt
 \begin{array}{ll}
  \mb i_k &= \sigma(\mb W_{\mb x\mb i}\mb x_k + \mb W_{\mb h\mb i}\mb h_{k-1} + \mb b_{\mb i}) \\
  \mb f_k &= \sigma(\mb W_{\mb x\mb f}\mb x_k + \mb W_{\mb h\mb f}\mb h_{k-1} + \mb b_{\mb f}) \\
  \mb c_k &= \mb f_k\mb c_{k-1} + \mb i_k\tanh(\mb W_{\mb x\mb c}\mb x_k + \mb W_{\mb h\mb c}\mb h_{k-1} + \mb b_{\mb c}) \\
  \mb o_k &= \sigma(\mb W_{\mb x\mb o}\mb x_k + \mb W_{\mb h\mb o}\mb h_{k-1} + \mb b_{\mb o}) \\
  \mb h_k &= \mb o_k\tanh(\mb c_k)
 \end{array}
 \right\},
 \label{eq:lstm}
\end{equation}
where $k$ indexes the timestep, $\mb i$, $\mb f$ and $\mb o$ are the input, forget and output gates, $\mb c$ and $\mb h$ are the cell and output states, all respectively, and peephole connections and some bias terms are omitted for simplicity.
By ignoring recurrent connections $k-1$ for the feed-forward case and making the LSTM completely ungated, i.e.~$\mb i = \mb f = \mb o = \mb I$, and initializing the cell state to the input $\mb c_{k-1} = \mb x$, we are left with a residual layer.
Again by extension then, cross-residual layers can be thought of as feed-forward, ungated LSTMs whose cell states are additively coupled.
LSTM forget gates then are analogous to cross-residual weight layers.
And indeed, this is much like highway networks' carry gate, since highway layers can be viewed as feed-forward LSTMs with only forget gates \cite{srivastava_2015}.
A major difference to note though is that cross-residual layers couple the transformed input $\mathcal{H}$ with \emph{multiple} and usually \emph{different} prior cell states $\mb c_{k-1}^{(t)}$ or information highways $\mb x^{(t)}$.

\vspace{1.6mm}
\noindent \textbf{Similarities to Ladder Networks} \cite{ramus_2015}\textbf{.}
Structurally, the building blocks of cross-residual learning bears some resemblance to the layout in ladder networks \cite{ramus_2015}.
In ladder networks, two encoders and one decoder joined via lateral connections are used to jointly optimize a weighted sum over a cross-entropy and reconstruction loss and have thus proven successful in semi-supervised learning.
As part of the reconstruction process, a Gaussian noise term is injected in one of the encoders and the decoder receives a combination of this noisy signal via a lateral connection and a vertical ``feedback'' connection to reconstruct the original input into the noisy encoder.
Since the mapping term $\mathcal{F}(\mb x)$ in residual learning can be viewed as perturbation term, albeit learned unlike in ladder networks, both models essentially are trying to fit the input subject to some additive nonlinear perturbation.
For cross-residual learning, although we use shortcut connections instead of lateral connections as in ladder networks, both designs operate on the principle that combining channels of information at the same structural level in the network can ultimately result in a model with higher learning capacity under less constraints, e.g.~for ladder networks, less labeled data requirements since it is semi-supervised.

\begin{figure*}[t]
  \centering
  \includegraphics[width=7.0in]{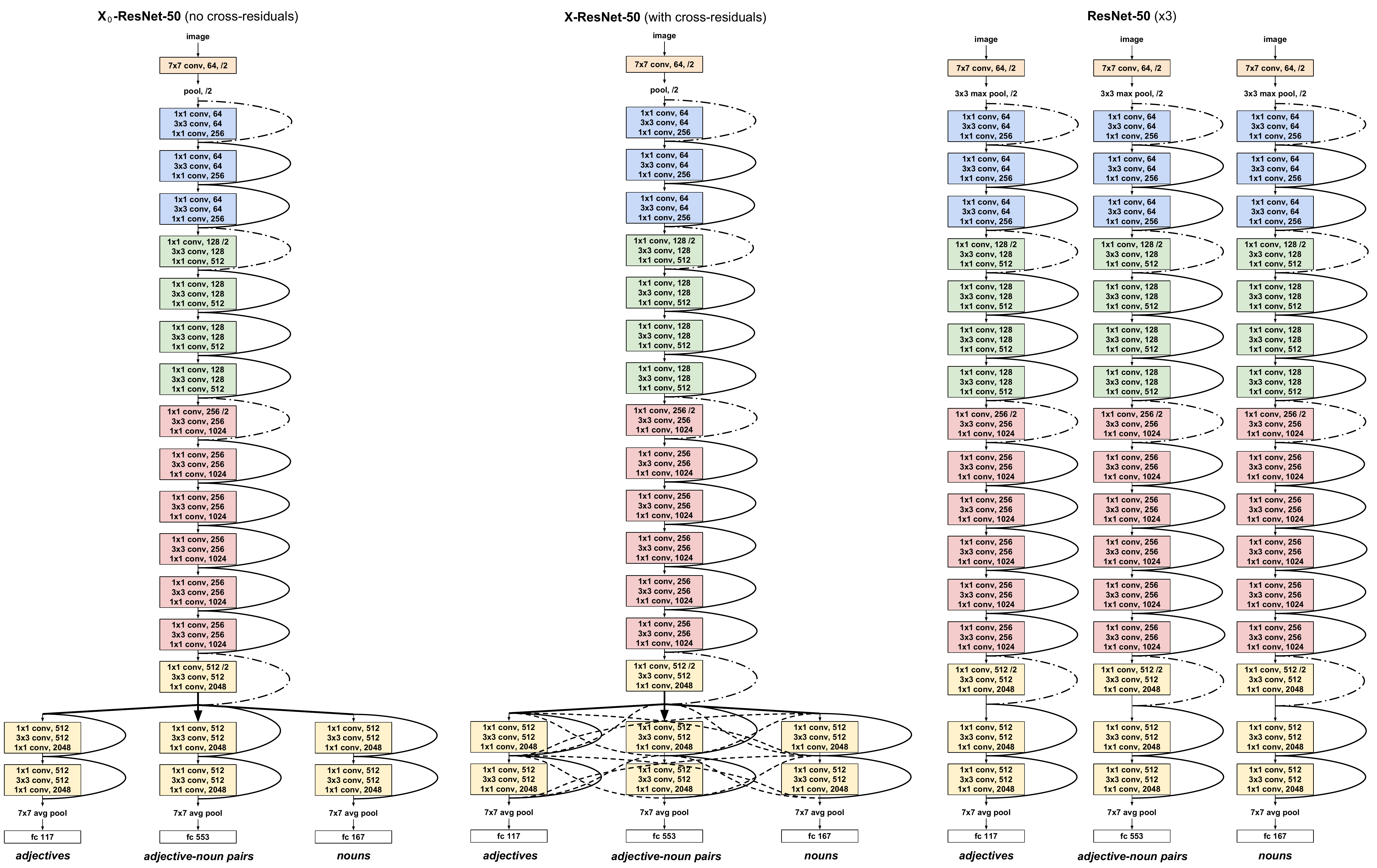}
  \caption{Example network architecture layouts for a standard multitask residual network, a multitask cross-residual network, and single-task residual networks, respectively, with 50 layers. Solid shortcuts (---) indicate identity, dash-dotted (-$\cdot$-$\cdot$-) shortcuts indicate $1\times1$ projections, and dashed (-$\,$-$\,$-) shortcuts indicate cross-residual weighted connections. Residual weight blocks show three convolutions grouped for space.}
  \label{fig:network_layouts}
\end{figure*}

\section{Multitask Cross Residual Nets}
\label{sec:mt_xresidual}

While there may be a number of settings that would benefit from cross-residual learning, we illustrate one natural setting here in multitask learning \cite{caruana_1997}.
To implement a multitask network, a common approach \cite{collobert_2008,ghifary_2015,liu_2015,ramsundar_2015} is to introduce a branching point in the architecture that leads to one network head per task, e.g.~see Figure \ref{fig:overview}.
In Table \ref{tab:mt_xresnet} and Figure \ref{fig:network_layouts}, we show 50-layer multitask residual networks with a branching point at the last input size reduction.
The earlier in the network this branching point is introduced the larger the input feature map size is to the individual network heads, often resulting in multitask networks with a large memory footprint.
On the other hand, if the branching point begins deeper in the network, the representational specialization available for each task is limited to a small space of high-level abstract features.
In our design of a multitask cross-residual network (X-ResNet), we address this latter problem by allowing additional cross-task mixing via cross-residual weights which cheaply increases late-layer representational power without requiring large input feature spaces.
While it is possible to completely forego a branching point in the network design and simply couple multiple network towers using cross-residual skip connections, this results in a composite network that is very memory intensive and only feasible in a multi-GPU environment (though this could be somewhat alleviated by freezing weights, e.g.~in combination with greedy layerwise training).

In addition, to introduce some task specialization, at the branching point in our multitask network design and before the cross-residual layers, we move the last ReLU activation and batch normalization canonically present inside the residual building block outside, placing it after the elementwise addition such that there is one per task.
This helps to produce a slightly different normalization for each task branch and in practice, slightly improves performance.
As in most multitask networks with a branching point, the total network loss is taken to be a combination of each of the individual network head losses.
While some tune the loss weight for each of these network heads, we simply use the unweighted sum over all the network head losses.

\begin{table}[t]
  \centering
  \begin{tabular}{C{1.23cm}|C{1.8cm}|C{1.8cm}|C{1.8cm}} \hline
    Output Size & Adjective & Adj-Noun Pair & Noun \\ \hline
    $112\times112$ & \multicolumn{3}{c}{$7\times7$, 60 /2} \\ \hline
    $56\times56$ & \multicolumn{3}{c}{$3\times3$ max pool /2} \\ \hline
    $56\times56$ & \multicolumn{3}{c}{$\begin{bmatrix} 1\times1, 64 \\ 3\times3, 64 \\ 1\times1, 256 \end{bmatrix}$$\times3$} \\ \hline
    $28\times28$ & \multicolumn{3}{c}{$\begin{bmatrix}1\times1, 128\\ 3\times3, 128\\ 1\times1, 512\end{bmatrix}$$\times4$} \\ \hline
    $14\times14$ & \multicolumn{3}{c}{$\begin{bmatrix}1\times1, 256\\ 3\times3, 256\\ 1\times1, 1024\end{bmatrix}$$\times6$} \\ \hline
    \multirow{2}{*}{$7\times7$} & $\begin{bmatrix}1\times1, 512\\ 3\times3, 512\\ 1\times1, 2048\end{bmatrix}$ & $\begin{bmatrix}1\times1, 512\\ 3\times3, 512\\ 1\times1, 2048\end{bmatrix}$ & $\begin{bmatrix}1\times1, 512\\ 3\times3, 512\\ 1\times1, 2048\end{bmatrix}$ \\
     & $\times3$ & $\times3$ & $\times3$ \\ \hline
    \multirow{3}{*}{$1\times1$} & avg pool & avg pool & avg pool \\ \cline{2-4}
     & 117-d fc & 553-d fc & 167-d fc \\ \cline{2-4}
     & softmax & softmax & softmax \\ \hline
  \end{tabular}
  \caption{Multitask Residual Network with 50 layers (without cross-residuals). Bracketed blocks are stacked residual building blocks. Downsampling is performed by stride 2 after stacked residual blocks.}
  \label{tab:mt_xresnet}
\end{table}

\section{Multitask Visual Sentiment}
\label{sec:mt_visual_sentiment}

In order to illustrate the utility and effectiveness of cross-residual layers when used in multitask networks, we re-frame visual sentiment concept detection in a multitask context.
In particular, we use the Visual Sentiment Ontology (VSO)\footnote{\url{https://visual-sentiment-ontology.appspot.com}} \cite{borth_2013} and cast affective mid-level concept detection as a multitask learning problem.
We chose the VSO dataset for our preliminary experiments because it presents a visual affect challenge currently of rising interest in the multimedia community as VSO has since had multilingual extensions \cite{jou_2015} and been applied in aesthetics understanding \cite{subh_2013}, emotion prediction \cite{jiang_2014,jou_2014}, popularity modeling \cite{khosla_2014}, and more.
While similar problems could have been created from other datasets, e.g.~CIFAR-100 where we might choose to predict classes and superclasses simultaneously, the adjective-noun pair detection problem can be recast to naturally fit the multitask setting with a sufficiently large accompanying image corpus over three tasks, i.e.~adjective, noun and adjective-noun pair, while other image datasets are often smaller and/or only consist of two learning tasks which yield a small number of task interactions.

Given the diversity of adjective-noun pairs, including concepts like `cute dress', `gentle smile', `scary skull', `wild rose' and `yummy cake', there is both a considerable amount of semantic variance in VSO as well as inter-class visual variance due to the image data being gathered from social media streams.
As a result, to cope with this diversity and variance, we believe that exploiting cross-task correlations as part of the network design is important, especially when the tasks are tightly related as they are with noun, adjective, and adjective-noun pair concept detection.

We additionally note that even though VSO \cite{borth_2013} argues that the noun component of the ANP serves to visually ground the mid-level concept, no experiments were actually ever run to determine the performance of detecting adjective (or even, noun) concepts separately\footnote{From independent communication with the authors.}.
Our evaluation thus also serves as the first evaluation on the VSO dataset to benchmark noun-only and adjective-only detection performance.

\subsection{Multitask-structured VSO}
\label{ssec:mt_vso}

Briefly, the data in VSO \cite{borth_2013} was originally collected from the social multimedia platform, Flickr\footnote{\url{https://www.flickr.com}}, using psychology-grounded seed queries from \emph{Plutchik's Wheel of Emotions} \cite{plutchik_1980} which consists of 24 basic emotions, such as \emph{joy}, \emph{terror}, and \emph{anticipation}.
The query results yielded images with user-entered image tags which were annotated using a part-of-speech tagger for identifying adjective and noun components and parsed for sentiment strength.
The identified adjective and noun components were combined, checked for semantic consistency and filtered based on sentiment strength then used to feed back as queries to Flickr to filter based on frequency of usage.
A subsampling of adjective-noun pair combinations is then done to prevent many adjective variations on any one noun, resulting in the final visual sentiment ontology.
The adjective-noun pairs were then used to query and pull down an image corpus from Flickr, limiting to at most 1,000 images per concept.

The image dataset in VSO \cite{borth_2013} has a long tail distribution where some adjective-noun pair concepts are singletons and do not share any adjectives or nouns with other concept pairs.
As a result, we use a subset of VSO and use it to perform adjective, noun, and ANP concept detection in social images, specifically, as a multitask learning problem.
The original VSO dataset \cite{borth_2013} consists of a refined set of 1,200 ANP concepts.
Since there are far less adjectives that serve to compose these adjective-noun pairs, and also some nouns that are massively over-represented in the ontology, we filtered to keep concepts that matched the following criteria:
(1) adjectives with $\geq$3 paired nouns,
(2) nouns that are not overwhelmingly biasing, v.s.~\emph{face} or \emph{flowers}, and non-abstract, unlike \emph{loss}, \emph{adventure} or \emph{history}, and
(3) ANPs with $\geq$500 images.
It is helpful to think of ANPs as a bipartite graph with nouns and adjectives on either side and valid ANPs as edges.
From these conditions, we obtained a visual sentiment sub-ontology, suitable for multitask learning, that normalized the number of adjective and noun nodes while ensuring maximal ANP edge coverage.
The final multitask-flavored VSO contains 167 nouns and 117 adjectives which form 553 adjective-noun pairs over 384,258 social images collected from Flickr.

\subsection{Experiments \& Discussion}
\label{ssec:experiments}

In our experiments, we use a 80/20 partition of the multitask VSO data stratified by adjective-noun pairs resulting in 307,185 images for training and 77,073 for test at 224$\times$224. 
All our residual layers use ``B option'' shortcut connections as detailed in \cite{he_2016} where projections are only used when matching dimensions (stride 2) and other shortcuts are identity.
Except for cross-residual weight layers, projections are performed with a $1\times1$ convolution with ReLU activation and batch normalization as in \cite{he_2016}.
For our cross-residual weight layers $\mb W_s^{(j)}$, we use the identity on self-shortcut connections $\mb W_s^{(t)}=\mb I$ and a cheap channelwise scaling layer for cross-task connections $\mb a\odot\mb x,\,\,\forall\,\,j\neq t$ which adds no more than 2,048 parameters each, i.e.~so in our case, after branching we have $\mb x \in \mathbb{R}^{7\times7\times2048}$ and so $\mb a \in \mathbb{R}^{1\times1\times2048}$ for scaling.

For training multitask networks, we initialized most layers using weights from a residual network (ResNet) model trained on ILSVRC-2015 \cite{imagenet}, but done such that for layers \emph{after} the branching point in our network we initialize them to the \emph{same} corresponding layer weights in the original ResNet model.
For cross-residual weight layers, we follow \cite{he_2016} and initialize them as in \cite{he_2015}, i.e.~zero mean random Gaussian with a $\sqrt{2/n_l}$ standard deviation where we set $n_l$ to be the average of input and output units layerwise.
No dropout \cite{srivastava_2014} was used in residual or cross-residual networks.
We use random flips of the input at training.
We trained our cross-residual networks with stochastic gradient descent using a batch size of 24, momentum of 0.9 and weight decay of 0.0001.
We used a starting fixed learning rate of 0.001 and decreased it by a factor of ten whenever the loss plateaued until convergence.
All networks and experiments were run using a single NVIDIA GeForce GTX Titan X GPU and implemented with Caffe \cite{jia_2014}.

We baseline against four single-task architectures: a variant of AlexNet \cite{krizhevsky_2012} swapping pooling and normalization layers called CaffeNet \cite{jia_2014}, the first iteration of the GoogLeNet architecture \cite{szegedy_2015} denoted as Inception-v1 which uses a bottlenecked $5\times5$ convolution in the sub-modules, the 16-layer version of VggNet \cite{simonyan_2015} (VggNet-16), and the ResNet architecture \cite{he_2016} with 50-layers (ResNet-50).
Each of these single-task architectures were fine-tuned from an ImageNet-trained model and represent competitive baselines that achieved top ranks in ILSVRC tasks in the past.
In addition, we also evaluated against DeepSentiBank \cite{chen_2014}, also an AlexNet-styled model trained on the full, unrestricted VSO data \cite{borth_2013} to detect 2,089 ANPs.
We did not retrain \cite{chen_2014} but rather re-evaluated their model on the subset of 553 ANP concepts we focus on here; however, since we do not know the train and test image splits that they used, the result provided for DeepSentiBank \cite{chen_2014} could still be an over-estimate.
In Figure \ref{fig:network_layouts} (rightmost), we show the learning and inference paradigm represented by these single-task architectures with residual networks (ResNet) used as an example.
Each of these baselines treat the adjective, noun and adjective-noun recognition tasks as independent targets.

\begin{table}[t]
  \centering
  \begin{tabular}{lcccc}
     & \textbf{Task} & \textbf{\#Params} & \textbf{Top-1} & \textbf{Top-5} \\ \hline
    \multirow{3}{*}{Chance} & Noun & -- & \,\,\,0.60 & \,\,\,2.96 \\
    & Adj & -- & \,\,\,0.86 & \,\,\,4.20 \\
    & ANP & -- & \,\,\,0.18 & \,\,\,0.90 \\ \hline
    DeepSentiBank \cite{chen_2014} & ANP & 65.43 & \,\,\,7.86 & 11.96 \\ \hline
    \multirow{3}{*}{CaffeNet \cite{jia_2014}} & Noun & 57.55 & 36.11 & 63.48 \\
    & Adj & 57.35 & 23.84 & 51.20 \\
    & ANP & 59.13 & 18.84 & 41.57 \\ \hline
    \multirow{3}{*}{Inception-v1 \cite{szegedy_2015}} & Noun & 10.82 & 39.93 & 67.98 \\
    & Adj & 10.66 & 26.32 & 55.57 \\
    & ANP & 12.00 & 20.48 & 45.01 \\ \hline
    \multirow{3}{*}{VggNet-16 \cite{simonyan_2015}} & Noun & 134.94 & 41.64 & 69.51 \\
    & Adj & 134.74 & 28.45 & 57.77 \\
    & ANP & 136.53 & 22.68 & 47.70 \\ \hline
    \multirow{3}{*}{ResNet-50 \cite{he_2016}} & Noun & 23.90 & 41.64 & 69.81 \\
    & Adj & 23.80 & 28.41 & 57.87 \\
    & ANP & 24.69 & 22.79 & 47.82 \\ \hline
    \multirow{3}{*}{\textbf{X$_{\mb 0}$-ResNet-50}} & Noun & \multirow{3}{*}{$\begin{pmatrix} \, \\ 43.16 \\ \, \end{pmatrix}$} & 40.06 & 68.06 \\
    & Adj & & 26.81 & 56.09 \\
    & ANP & & 20.74 & 45.46 \\ \hline
    \multirow{3}{*}{\textbf{X$_{\mb I}$-ResNet-50}} & Noun & \multirow{3}{*}{$\begin{pmatrix} \, \\ 43.16 \\ \, \end{pmatrix}$} & 28.61 & 56.52 \\
    & Adj & & 17.98 & 43.10 \\
    & ANP & & 12.56 & 31.49 \\ \hline
    \multirow{3}{*}{\textbf{X$_{\mb s}$-ResNet-50}} & Noun & \multirow{3}{*}{$\begin{pmatrix} \, \\ 43.18 \\ \, \end{pmatrix}$} & \textbf{42.18} & \textbf{70.04} \\
    & Adj & & \textbf{28.88} & \textbf{58.50} \\
    & ANP & & \textbf{22.89} & \textbf{48.54} \\ \hline
  \end{tabular}
  \caption{Number of Parameters (millions) and Top-$k$ Accuracy (\%) on the Multitask VSO dataset. Note that X-ResNet-50 are multitask networks so classifiers are trained jointly in a single network while other methods train one specialized network per classification task.}
  \label{tab:mtvso_classifiers}
\end{table}

We summarize network parameter costs and top-$k$ accuracy on the multitask VSO tasks in Table \ref{tab:mtvso_classifiers}.
For network parameter costs, note that for Inception-v1 \cite{szegedy_2015} we did not count the parameters from auxiliary heads although they are used during training.
Top-$k$ accuracy denotes the percentage of correct predictions within the top $k$ ranked decision outputs.

\subsubsection{Adjective \texttt{vs.}~Noun \texttt{vs.}~ANP Detection}

In general, as originally posited by the VSO work \cite{borth_2013}, in terms of problem difficulty ordering, noun prediction is indeed ``easier'' as visual recognition task than adjective prediction.
However, though not in stark contrast to \cite{borth_2013}, and although there are indeed more ANP classes than nouns and adjectives, we still did expect to observe higher accuracy rates for ANP concept detection than we did, expecting that the rates would be much closer to that of noun detection and not lower than adjective detection since \cite{borth_2013} argues that adjectives lack visual grounding.
We suspect that this difference by almost a half at top-1 between noun and ANP detection may point to the difficulty of the ANP detection problem in a slightly different sense than difficulty for the adjective detection problem.
For adjective detection, visual recognition difficulty is likely to arise from visual variance, e.g.~there may be a wide range of visual features required to describe the concept `pretty'.
However, for ANP detection, we believe that visual recognition difficulty is more likely due to visual nuances than overall visual variance.
Much like fine-grained classification, this may imply that in ANP concept detection, concepts like `sad dog' and `happy dog' may share many visual characteristics but differ on few but highly distinguishing traits.
The hope then is that by using a scaling layer, which acts as a soft gating mechanism in cross-residual connections, these few but distinguishable characteristics are accentuated.

\subsubsection{Effects of Cross-residual Weighting}

In Table \ref{tab:mtvso_classifiers}, we also show results for multitask cross-residual networks with different types of weighting: no cross-residual weighting (X$_{\mb 0}$-ResNet-50), with all identity cross-residual weights (X$_{\mb I}$-ResNet-50) and with identity on the self-task connections and channelwise scaling on just cross-residuals as described earlier (X$_{\mb s}$-ResNet-50).
The multitask cross-residual networks without and with cross-residuals are illustrated in Figure \ref{fig:network_layouts} (leftmost and center, respectively), and all of these multitask networks use a residual network with 50-layers (ResNet-50) as the basis and branch as described in Section \ref{sec:mt_xresidual}.
As we might expect, when \emph{all} cross-residual weights are identity (X$_{\mb I}$-ResNet-50), the accuracy of the multitask network across all tasks drastically reduces since the ``amount'' of cross-task mixing is forced to be equally weighted.
Even as related as tasks might be forcing cross-residual weights equal across all tasks makes it difficult during learning for any single task to specialize and determine discriminative patterns useful for that specific task.
It may be tempting to then assume that the other extreme of making the cross-residual weights zero where $\mb W_s^{(j)} = \mb 0, \, \forall \, j \neq t$, i.e.~equivalent to a multitask network without cross-residuals (X$_{\mb 0}$-ResNet-50), allows more specialization and would naturally achieve the best discriminative performance.
However, we found that this actually consistently achieves lower accuracy across all tasks compared to its single-task equivalents (ResNet-50), e.g.~$\sim$9\% worse relative on ANP detection.
We hypothesize that without cross-residuals the performance case becomes upper-bounded by the shared representation learned before the branching the multitask network.

Once we allow for even some simple learned weighting on the cross-residuals, like a channelwise scaling (X$_{\mb s}$-ResNet-50), the predictive performance of the multitask network improves, outperforming both the case when no cross-residuals are used as well as equally weighted cross-residuals.
In general, we observed that multitask networks achieved comparable performance to the three specialized single-task networks with just a single network while requiring less than 60\% of the combined parameters of the three single-task networks ($\sim$43.2M vs.~$\sim$72.4M).
This confirms our original hypothesis that the low-level representations can be shared across these related tasks and can be generalized to perform well across all tasks.
However, in order to ensure that we do not take a hit in accuracy by generalizing, weighted cross-residuals layers can be used which, at a very marginal parameter cost, enable the multitask network to match the performance of specialized single-task networks.
Notably, as we had hoped, the highest gain from using cross-residuals was on the most difficult of the three tasks: ANP detection.
We observe that adding scaling cross-residual weights improves the concept detection performance by as much as $\sim$10.37\% relative on the ANP detection task compared to without any weighting.

Though we do not claim that our cross-residual multitask network (X$_{\mb s}$-ResNet-50) definitively achieves a significantly higher accuracy over the single-task networks, we do note that we observed marginally better concept detection rates with our network across all tasks.
Since we only used two cross-residual layers in our multitask network (c.f.~Figure \ref{fig:network_layouts}), it is possible that increasing the number of stacked cross-residual layers or beginning the branching in the network earlier could improve the overall cross-task performance; however, doing so would naturally come at increased parameter cost.
Nonetheless, we believe that all of these observations show that jointly learning across related tasks with cross-task information mixing even at the late layers of a network can simultaneously improve the network's capacity to discriminate and generalize.

\begin{figure}[t]
  \centering
  \includegraphics[width=3.31in]{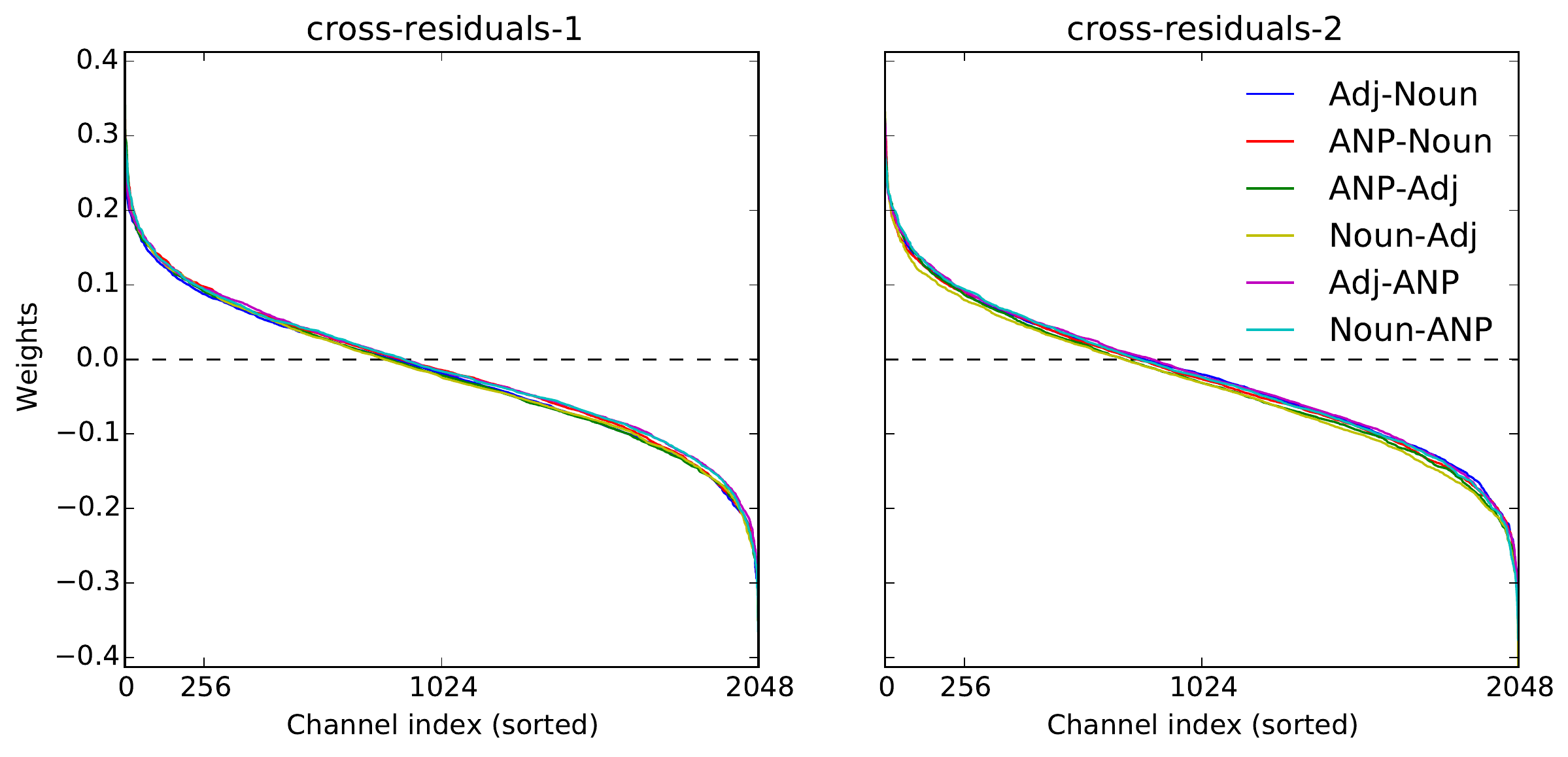}
  \caption{Example learned unnormalized cross-residual weights (sorted). Legend notation refer to cross-residual connections as \texttt{SourceTask}-\texttt{TargetTask}. Left and right plots show cross-residual weights of the first and second (feed-forward direction) cross-residual layers as in Figure \ref{fig:network_layouts} (center), respectively.} 
  \label{fig:xresidual_weights}
\end{figure}

To further reinforce that the optimal weightings for cross-residual connections are unlikely to be zero or identity, in Figure \ref{fig:xresidual_weights}, we show the unnormalized weight magnitudes of a learned multitask cross-residual network sorted by channel index for two cross-residual layers in a network structured as in Figure \ref{fig:network_layouts} (center).
If an all zero or identity cross-residual connection were to be optimal, we would expect to see a plateau with many weights near zero or one.
Instead, we observe that mostly non-negative cross-task weights were learned across all shortcut connections such that the overall network objective was optimized.
Additionally, we note that though the weight magnitudes are indeed small, this also follows from intuition in the original residual network work \cite{he_2016} that these small, but non-zero weights are precisely what enable residual networks to be made very deep.

\subsubsection{Example Multitask Detection Results}

\begin{figure}[t]
  \centering
  \includegraphics[width=3.22in]{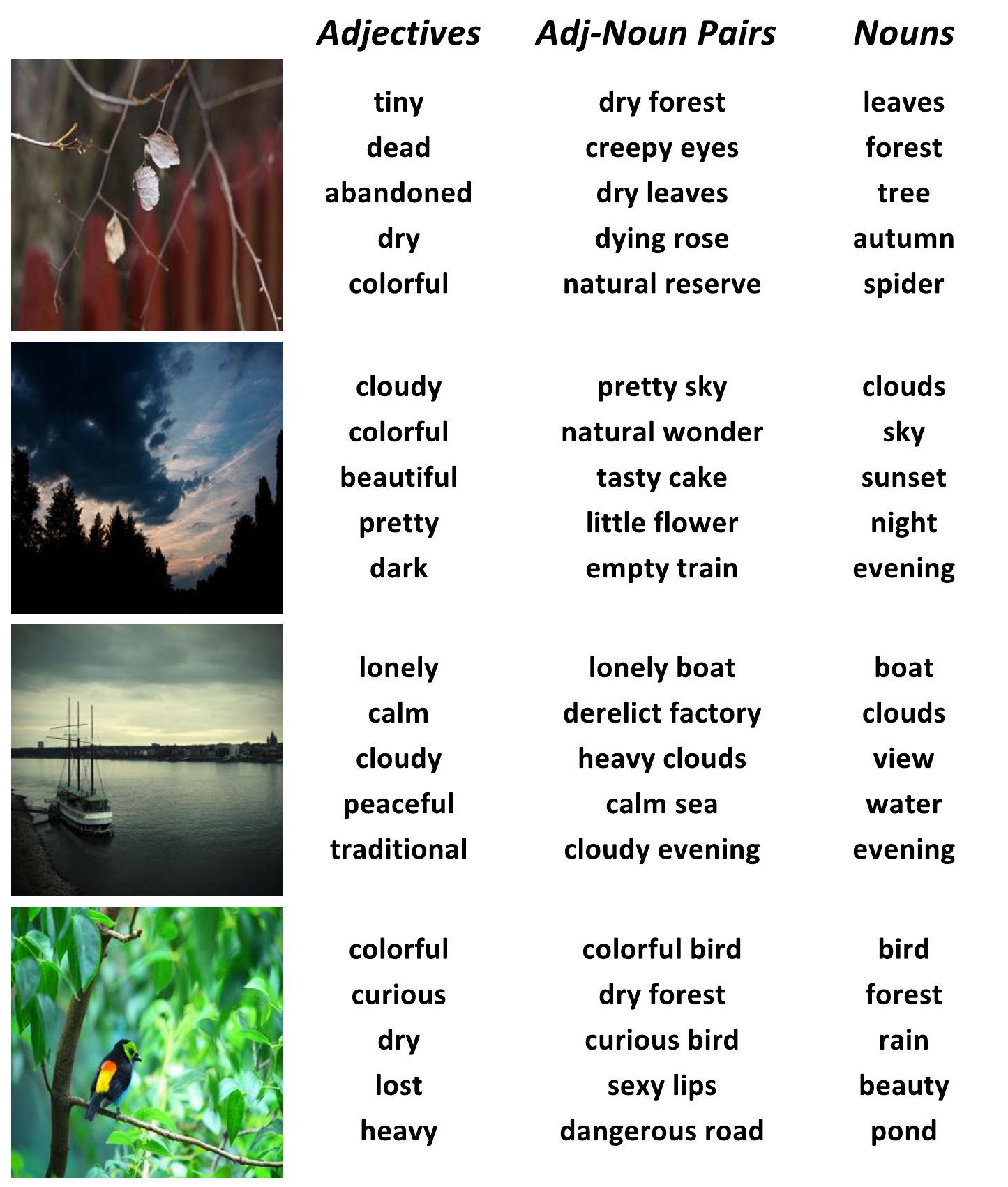} 
  \caption{Example top-5 classification results of adjective, noun, and adjective-noun pair concepts using our multitask cross-residual network.}
  \label{fig:detection_results}
\end{figure}

In Figure \ref{fig:detection_results}, we show example classification results from our multitask cross-residual network.
Note the presence of both intra- and inter-relatedness between tasks in the top detected concepts.
In many cases, the cross-residual network is able to surface concepts not visually present but intuitively related; for example, in the first image, `spider' is a detected noun which may be a result of either the branches in the image or the visual co-occurrence of the `spider' concept in the training set with other top ranked concepts like `tiny' (adjective) and `leaves' (noun).
As a potential failure case, in the last image, the ANP `sexy lips' was ranked highly possibly due to relatedness learned with the `colorful' adjective concept.
In these cases, just as with over regularization in other learning settings, the network may have indeed have learned a more general representation, but as a result is unable to decouple certain learned relationships.
Such cases may be easily addressed in cross-residual networks by giving cross-task weighting layers more computational budget, e.g.~convolutional projections, to model more complicated task relationships.
Overall, we observe here that the multitask cross-residual network is able to successfully co-detect concepts across multiple related visual recognition tasks.


\section{Conclusions \& Future Work}
\label{sec:conclusion}

We presented an extension of residual learning enabling information mixing between related tasks called \emph{cross-residual learning} (CRL) achieved by coupling the residual to other related tasks to ensure the learned mapping is not too far from other task representations.
This enables more generalized representations to be learned in a deep network that are useful for multiple related tasks while preserving their discriminative power.
We also showed how cross-residuals can be used for multitask learning by integrating cross-residual layers in a fan-out multitask network.
We illustrated how such a multitask cross-residual network can achieve competitive, or even better, predictive performance on a visual sentiment concept detection problem as compared to specialized single-task networks but with $>$40\% less parameters, while also outperforming a standard multitask residual network with no cross-residuals by about 10.4\% relative on adjective-noun pair detection, the hardest of the three related target tasks.
Without cross-residual connections, we observed a $\sim$9\% drop in accuracy on ANP detection, indicating the importance of using cross-residuals.
In addition, we showed the importance of cross-residual weighting over simply forcing identity cross-residual connections since equally weighting cross-task connections bottlenecks the information flow in the network.

We believe cross-residual networks are also applicable to other learning settings and domains, and can be extended in several ways.
Cross-residual networks can be applied to other multitask learning settings where we are not only interested in classification but also other tasks like reconstruction, object segmentation, etc.
Likewise, cross-residual networks are likely to be useful in domain transfer and adaptation problems where, for example, network tower weights are frozen but cross-residual weights are learned.
Architecturally, while we only explored the canonical shortcut connections of \cite{he_2016} and used a channelwise scaling layer for the cross-residual, there is recent work exploring different types of transforms and gating on shortcuts \cite{he_2016_imap} that can also be extended to the self- and cross-connections in cross-residual networks.
We plan to explore these learning settings and network architectures in the future.

\section*{Acknowledgements}
We thank our reviewers for their helpful and constructive feedback.
We also thank Rogerio Feris for insightful discussions on the network design and Tao Chen for support with the Visual Sentiment Ontology (VSO) dataset as well as discussions on prior VSO experiments.

\bibliographystyle{abbrv}

%

\end{document}